\documentclass[letterpaper, 10 pt, conference]{IEEEtran}
\usepackage{times}
\usepackage{cite}
\usepackage{multicol}
\usepackage{amsmath}
\usepackage{amsfonts}
\usepackage{mathtools}
\usepackage{booktabs}
\usepackage{graphicx}
\usepackage{enumerate}
\usepackage{dashrule}
\usepackage{xcolor}
\usepackage{algpseudocode}
\usepackage[linesnumbered, ruled, vlined]{algorithm2e}
\usepackage[hidelinks]{hyperref}
\usepackage{amssymb}

\newcommand{\vct}[1]{\mathbf{#1}}         
\newcommand{\mtr}[1]{\mathbf{#1}}         
\newcommand{\bzero}{\mathbf{0}}           
\newcommand{\vzero}{\mathbf{0}}           

\newcommand{\transpose}{\intercal}

\newcommand{\norm}[1]{\left\lVert #1 \right\rVert}

\newcommand{\calP}{\mathcal{P}}

\newcommand{\R}{\mathbb{R}}

\newcommand{\matQ}{\mtr{Q}}
\newcommand{\vecC}{\vct{c}}
\newcommand{\matA}{\mtr{A}}
\newcommand{\vecB}{\vct{b}}
\newcommand{\matG}{\mtr{G}}
\newcommand{\vecH}{\vct{h}}

\newcommand{\vx}{\vct{x}}
\newcommand{\vs}{\vct{s}}
\newcommand{\vxi}{\boldsymbol{\xi}}

\newcommand{\feasSet}{\mathcal{F}}
\newcommand{\dualSet}{\mathcal{D}}

\newcommand{\ineqSet}{\mathcal{I}}
\newcommand{\convexSet}{\mathcal{C}}
\newcommand{\proj}[1]{\Pi_{#1}}

\newcommand{\vse}{\vct{s}_e}
\newcommand{\vsi}{\vct{s}_i}

\newcommand{\vy}{\vct{y}}
\newcommand{\vz}{\vct{z}}
\newcommand{\vw}{\vct{w}}

\newcommand{\U}{\vct{U}}          
\newcommand{\Ut}{\tilde{\U}}      

\newcommand{\vr}{\vct{r}}

\newcommand{\vncpFun}[1]{\boldsymbol{\phi}^{#1}}

\newcommand{\ncpJac}[1]{\boldsymbol{\Phi}_{#1}}
\newcommand{\pertncpJac}[1]{\tilde{\boldsymbol{\Phi}}_{#1}}

\newcommand{\rhod}{\rho_{d}}      
\newcommand{\rhoe}{\rho_{e}}      
\newcommand{\rhoi}{\rho_{i}}      
\newcommand{\rhon}{\rho_{n}}      

\newcommand{\gamme}{\gamma_{e}}    
\newcommand{\gammi}{\gamma_{i}}  

\newcommand{\xE}{\vx_{E}}
\newcommand{\yE}{\vy_{E}}
\newcommand{\zE}{\vz_{E}}
\newcommand{\sE}{\vs_{E}}
\newcommand{\wE}{\vw_{E}}

\newcommand{\Qtil}{\tilde{\matQ}}

\newcommand{\Id}{\mtr{I}}

\newcommand{\dx}{\Delta\vx}
\newcommand{\dy}{\Delta\vy}
\newcommand{\dz}{\Delta\vz}
\newcommand{\ds}{\Delta\vs}

\newcommand{\vge}{\mathbf g_e}
\newcommand{\vgi}{\mathbf g_i}

\newcommand{\vtheta}{\boldsymbol{\theta}}

\usepackage{cleveref}
\Crefname{definition}{Definition}{Definitions}
\Crefname{proposition}{Proposition}{Propositions}
\Crefname{theorem}{Theorem}{Theorems}
\Crefname{figure}{Fig.}{Figs.}
\Crefname{table}{Table}{Tables}
\Crefname{equation}{Eq.}{Eqs.}
\Crefname{section}{Section}{Sections}
\Crefname{subsection}{Section}{Sections}
\Crefname{subsubsection}{Section}{Sections}
\Crefname{algorithm}{Algorithm}{Algorithms}

\usepackage{glossaries}
\newacronym{qp}{QP}{quadratic programming}
\newacronym{ai}{AI}{artificial intelligence}
\newacronym{mpc}{MPC}{model predictive control}
\newacronym{sqp}{SQP}{sequential quadratic programming}
\newacronym{kkt}{KKT}{Karush--Kuhn--Tucker}
\newacronym{alm}{ALM}{augmented Lagrangian method}
\newacronym{ipm}{IPM}{interior-point method}
\newacronym{nipm}{NIPM}{non-interior-point method}
\newacronym{ecj}{ECJ}{extended conservative Jacobian}
\newacronym{ncp}{NCP}{nonlinear complementarity problem}
\newacronym{lcp}{LCP}{linear complementarity problem}
\newacronym{nip}{NIP}{non-interior point}
\newacronym{ift}{IFT}{implicit function theorem}

\newcommand{\orcid}[1]{\href{https://orcid.org/#1}{\includegraphics[width=0.6em]{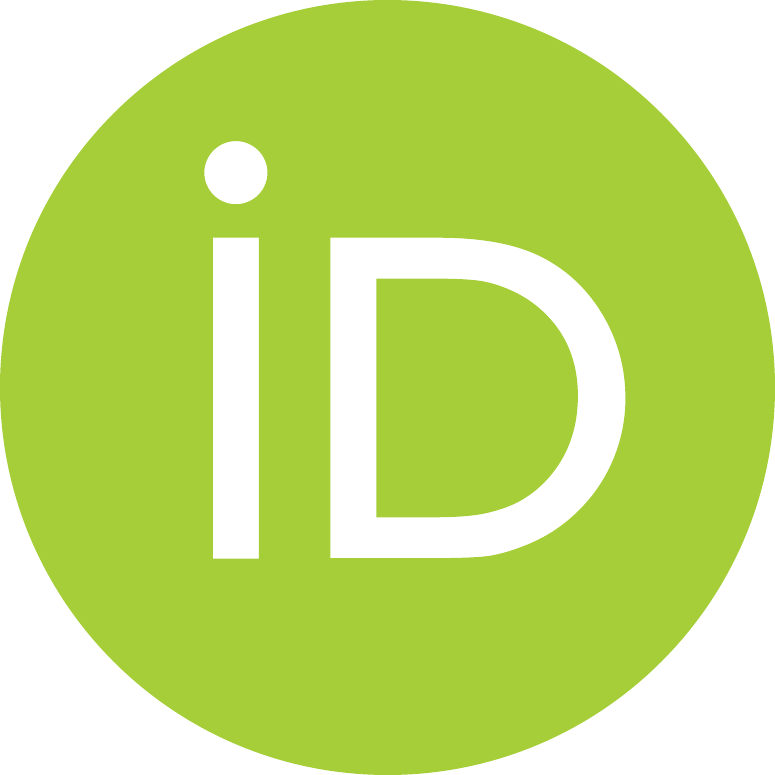}}}

\newcommand{\reb}[1]{{\color{black}#1}}
\newif\ifpublished

\newcommand{\published}[1]{%
  \ifnum\pdfstrcmp{#1}{True}=0\relax
    \publishedtrue
  \else
    \publishedfalse
  \fi
}

\published{True} 

\makeatletter
\def\blfootnote{\gdef\@thefnmark{}\@footnotetext}
\makeatother
\IEEEoverridecommandlockouts
\begin{document}

\bstctlcite{IEEEexample:BSTcontrol}

\title{\textsc{Elastic Odyn}: Differentiable Optimization for Infeasible Control and Learning in Robotics}

\ifpublished
\author{Aristotelis Papatheodorou$^{1}$\orcid{0000-0003-0290-7071}\quad 
Jose Rojas$^{2}$\orcid{0000-0001-5580-6616}\quad 
Ioannis Havoutis$^{1}$\orcid{0000-0002-4371-4623}\quad 
Carlos Mastalli$^{2}\orcid{0000-0002-0725-4279}$%
\thanks{\scriptsize $^{1}$A.~Papatheodorou and I.~Havoutis are with the Dept.\ of Engineering Science, Univ.\ of Oxford, UK. $^{2}$J.~Rojas and C.~Mastalli are with the RoMI Lab, Heriot-Watt Univ., Edinburgh, UK. A.P. is supported by the Clarendon Fund and the JPMorgan Chase AI PhD Fellowship. J.R. by the AMAN project, funded by Tata Consultancy Services.}
\thanks{\scriptsize$^{*}$Open-source implementation will be released upon acceptance.}
}
\else
\author{Author Names Omitted for Anonymous Review.
\thanks{$^{*}$Open-source implementation will be released upon acceptance.}
}
\fi

\maketitle

\thispagestyle{empty}
\pagestyle{empty}

\begin{abstract}
Robotic systems routinely encounter conflicting objectives, modeling errors, and degenerate contact conditions that render quadratic programs (QPs) infeasible. Yet most optimization solvers and differentiable QP layers assume feasibility, leading to numerical failures, unstable gradients, or solver breakdown when constraints cannot be simultaneously satisfied. We present \textsc{Elastic Odyn}, a primal--dual non-interior-point QP solver that handles infeasibility through smooth squared-$\ell_2$ elastic relaxations. The formulation remains well posed under ill-conditioning and degeneracy, supports warm starting, and converges to closest-to-feasible solutions, with lightweight refinement recovering physically meaningful dual variables. Building on this framework, we develop \textsc{Elastic OdynLayer}, a differentiable QP layer with stable gradients under infeasibility, and \textsc{Elastic OdynSQP}, an SQP method that resolves inconsistent subproblems and intrinsically infeasible optimal control tasks through selective constraint \reb{elasticity}. Across benchmark QPs, singular contact mechanics, differentiable parameter identification, and quadrupedal and humanoid trajectory optimization, \textsc{Elastic Odyn} outperforms state-of-the-art elastic QP solvers in robustness, warm-start performance, and convergence reliability, enabling optimization, simulation, control, and learning beyond standard feasibility assumptions.
\end{abstract}
\vspace{0.3em}
\vspace{-0.3em}


\section{Introduction}\label{sec:intro}
\Gls{qp} is a fundamental tool across science and engineering.
In robotics and~\gls{ai}~\cite{diehl_robotics}\reb{,} optimization problems are often large-scale and numerically stiff, placing stringent requirements on solver robustness, scalability, and numerical stability.
Despite its central role, there is currently no~\gls{qp} solver that is simultaneously robust to infeasibility, numerically stable for complex problems, and compatible with differentiable optimization.

Robotics provides a particularly demanding setting.
Robots must plan their actions in dynamic and uncertain environments while enforcing safety-critical constraints.
Decisions are made in real time and often over long horizons, leading to high-dimensional optimization problems with several constraints~\cite{papatheodorou_24}.
In practice, feasibility is frequently violated due to modeling errors, external disturbances, or conflicting task requirements.
As a result, the quadratic subproblems arising in control pipelines may themselves become infeasible.

\Gls{mpc}~\cite{mastalli2022agile} computes control actions in a principled way, handling dynamics and constraints by repeatedly solving structured quadratic programs.
However, \gls{mpc} scales poorly and becomes increasingly ill-conditioned with longer prediction horizons.
Moreover,~\gls{mpc} solvers are not designed to handle \emph{infeasible constraint sets} gracefully, which can lead to failure or unpredictable behavior in practice~\cite{jallet2022alddp}.
Within~\gls{sqp} methods, infeasibility is typically
addressed through restoration stages~\cite{nocedal-optbook}.
In contrast, elastic mechanisms provide a unified and
computationally efficient means of recovering feasibility while resolving the problem.

Differentiable optimization has emerged as a promising paradigm for learning-based methods~\reb{\cite{amos2017optnet}}, where the forward pass solves a constrained problem and the backward pass propagates loss gradients through its solution. These gradients are governed by the~\gls{kkt} system, which couples primal and dual variables and enforces stationarity and complementarity, but not primal feasibility, since intermediate gradient-based iterates may be infeasible. As a result, differentiable optimization layers are highly sensitive to infeasibility and ill-conditioning, exposing a fundamental deficiency of standard differentiable~\gls{qp} frameworks.

\subsection{Contributions}\label{sec:contributions}

This paper introduces \textsc{Elastic Odyn}$^*$, a framework for robust and differentiable quadratic programming under infeasibility. Our main contributions are:
\reb{
\begin{enumerate}[(i)]
\item \textbf{Elastic \textsc{Odyn}.}
We integrate the classical $\ell_2$ elastic relaxation into \textsc{Odyn}~\cite{rojas_odyn_2026}, introducing violations as \textbf{diagonal shifts rather than penalty curvature} to preserve conditioning and warm startability, while supporting both relaxed and hard constraints in the same QP.
\item \textbf{Dual recovery.}
We shift the relaxed constraints by their elastic violations and warm-start the resulting hard \gls{qp}, recovering physically interpretable duals in only a few iterations.
\item \textbf{Differentiable optimization under infeasibility.}
\textsc{Elastic OdynLayer} differentiates through the elastic KKT conditions, yielding well-defined gradients for feasible and infeasible \glspl{qp}, where feasibility-assuming layers are undefined.
\item \textbf{Infeasibility-aware SQP.}
\textsc{Elastic OdynSQP} uses constraint relaxation within SQP to handle inconsistent subproblems and nonlinear optimal control problems that are intrinsically infeasible.
\end{enumerate}
}
Unifying infeasibility handling, numerical robustness, and differentiability, \textsc{Elastic Odyn} opens new possibilities for optimization-based robotics and~\gls{ai} pipelines.

\section{Related Work}\label{sec:related_work}
Quadratic programming has been extensively studied, giving rise to a wide range of solvers with different trade-offs in scalability, robustness, and differentiability.
We review the main classes of optimization approaches, with a focus on infeasibility handling and differentiable structure.
\subsection{Optimization approaches and available solvers}

Active-set methods are among the earliest approaches to quadratic programming~\cite{goldfarb1983numerically}. Solvers such as~\textsc{qpOASES}~\cite{ferreau2014_qpoases} and the~\gls{sqp} solver~\textsc{Snopt}~\cite{gill-siam05} maintain the active constraint set and solve a sequence of equality-constrained subproblems. They are effective at small to medium scale and warm-startable, but scale poorly as constraint counts grow, due to exponential active-set combinatorics. Discrete active-set changes also induce nonsmooth solution mappings, \reb{leading to subgradients that hinders} training.

\Glspl{alm} handle constraints via penalty terms and iterative dual updates, exhibiting strong robustness to rank-deficient equalities and ill-conditioning. Solvers such as \textsc{Lancelot}~\cite{lancelot}, \textsc{Osqp}~\cite{osqp}, and \textsc{ProxQP}~\cite{proxqp} extend this framework to inequality-constrained problems, typically enforcing inequalities through projection or proximal operators. However, these strategies introduce nonsmoothness detrimental to numerical robustness and gradient-based training. Despite strong theoretical guarantees, reliance on penalty tuning and repeated inner solves limits ALM efficiency in real-time and learning-based settings.

\Glspl{ipm} dominate large-scale convex and nonlinear optimization. Solvers such as \textsc{Ipopt}~\cite{ipopt}, \textsc{Gurobi}~\cite{gurobi}, and \textsc{Piqp}~\cite{piqp} use barrier functions and primal–dual~\gls{kkt} systems to handle inequalities without explicit active-set enumeration, scaling well in feasible regimes. However, IPMs follow a central path that is well defined only when the problem is strictly feasible, which makes warm starting difficult and causes failures or numerical instabilities when problems become infeasible or nearly infeasible.

Motivated by these limitations,~\glspl{nipm}~\cite{Liao_McPherson_2019,rojas_odyn_2026} have emerged as alternatives that remain well defined outside the feasible set. Generally, they can be categorized into two families. \emph{Semismooth} methods such as~\textsc{FBstab}~\cite{liao2020fbstab} reformulate the~\gls{kkt} conditions via~\gls{ncp} functions like Fischer--Burmeister~\cite{qi1993nonsmooth}, with applications in contact simulation and control~\cite{todorov2010implicit}. \emph{Path-following} methods instead track smoothed~\gls{kkt} systems as the smoothing parameter vanishes, operating robustly under infeasibility~\cite{zhang2023iprqp} and yielding regular solution mappings well suited to implicit differentiation. Yet no existing method simultaneously addresses infeasibility and robustness within a unified differentiable framework.

\subsection{Differentiable optimization and limitations}
\reb{An important application of differentiable optimization is the training of optimization layers within learning pipelines. This requires reverse-mode gradients of the solution map. \textsc{OptNet}~\cite{amos2017optnet} differentiates the~\gls{kkt} conditions of quadratic programs, and \textsc{cvxpylayers}~\cite{cvxpylayers2019} extends this to convex programs. Both inherit the feasibility requirements and numerical sensitivity of~\glspl{ipm}. \textsc{QPlayer}~\cite{bambade-2024qplayer}, built on \textsc{ProxQP}~\cite{proxqp}, handles infeasibility through an elastic~\gls{alm}, but requires~\gls{ecj} differentiation~\cite{bolte-2020ecj} and additional relaxation variables in the backward pass. \textsc{BPQP}~\cite{pan2024bpqp} formulates the backward pass as a decoupled~\gls{qp}, \textsc{Alt-Diff}~\cite{sun2023altdiff} differentiates alternating primal--dual updates, and \textsc{DiffOpt.jl}~\cite{besancon2023diffopt} uses model transformations. All assume a feasible optimum.
The survey~\cite{kotary2021survey} highlights solution existence and uniqueness as common assumptions in differentiable optimization, while identifying constraint parametrization and feasibility guarantees as open challenges addressed here. Tracy and Manchester~\cite{Tracy2024OnTD} instead smooth primal--dual~\gls{ipm} solutions through central path relaxation, retaining barrier-parameter sensitivity but permitting only limited warm starting. Consequently, the resulting gradient reflect \emph{barrier smoothing} rather than true constrained sensitivities.

\textsc{Elastic Odyn} instead uses \emph{barrier and penalty functions} to handle degeneracy, ill conditioning, and primal infeasibility while supporting warm starting. Its differentiable counterpart, \textsc{Elastic OdynLayer}, is, to our knowledge, the first~\gls{qp} layer requiring no projections whose derivatives remain defined under infeasibility through direct differentiation of the elastic KKT system rather than a smoothed or lifted surrogate.}
\vspace{-5mm}
\section{Infeasibility in Quadratic Programming}
We consider convex quadratic programs of the form
\begin{equation}
\label{eq:standard_qp}
\begin{aligned}
\min\limits_{\vx} \quad
    & \tfrac{1}{2}\,\vx^\transpose \matQ\,\vx + \vecC^\transpose \vx \quad \text{s.t.} \quad \matA\,\vx = \vecB,\quad  \matG\,\vx \le \vecH,
\end{aligned}
\vspace{-2mm}
\end{equation}
with primal variables $\vx\in\R^n$, symmetric positive semidefinite cost matrix $\matQ\in\R^{n\times n}$, linear cost $\vecC\in\R^n$, and constraints defined by $\matA\in\R^{m\times n}, \vecB\in\R^m$ for the equalities and $\matG\in\R^{p\times n}, \vecH\in\R^p$ for the inequalities. Under standard regularity assumptions such as Slater's condition, optimal solutions of~\Cref{eq:standard_qp} are characterized by the~\gls{kkt} conditions, derived from the Lagrangian
\vspace{-1mm}
\begin{equation}
\label{eq:lagrangian}
\mathcal{L}(\vx,\vy,\vz)
=
\tfrac{1}{2}\,\vx^\transpose \matQ\,\vx + \vecC^\transpose \vx
+ \vy^\transpose(\matA\vx-\vecB)
+ \vz^\transpose(\matG\vx-\vecH),
\vspace{-1mm}
\end{equation}
where $\vy\in\R^m$ and $\vz\in\R^p$ are the multipliers associated with the equality and inequality constraints. Combining primal and dual feasibility, complementary slackness, and stationarity, the~\gls{kkt} conditions admit the equivalent variational form
\vspace{-2mm}
\begin{equation}
\label{eq:kkt_inclusion_intro}
\vzero \in \matQ\vx + \vecC + N_{\feasSet}(\vx),
\end{equation}
with feasible set $\feasSet \coloneq \{\vx\in\R^n \mid \matA\vx=\vecB,\; \matG\vx\leq\vecH\}$ and normal cone $N_{\feasSet}(\vx)$. This inclusion provides a unified representation that underpins the notions of primal and dual feasibility developed below.

\subsection{Primal infeasibility}
A point $\vx$ is primal infeasible if it violates at least one constraint, i.e., $\vx\notin\feasSet$. The geometric infeasibility is captured by $\mathrm{dist}(\vx, \feasSet) = \inf_{\vx'\in\feasSet} \| \vx - \vx'\|$, but is generally intractable to evaluate. We instead use the computable \emph{constraint violation}
\begin{equation}\label{eq:infeas_dist}
\mathrm{viol}(\vx) = \max\bigl(\|\matA\vx - \vecB\|,\ \|\vx - \proj\ineqSet(\vx)\|\bigr),
\end{equation}
where $\|\cdot\|$ is the Euclidean norm, $\ineqSet = \{\vx\in\R^n \mid \matG\vx \leq \vecH\}$, and $\proj\ineqSet(\vx)$ denotes the Euclidean projection onto $\ineqSet$. This quantity vanishes if and only if $\vx\in\feasSet$ and provides a tractable surrogate for $\mathrm{dist}(\vx, \feasSet)$.
For $\vx\notin\feasSet$, the variational inclusion is not satisfied, meaning that the~\gls{kkt} inclusion cannot hold, $\vzero \notin \matQ\vx + \vecC + N_{\feasSet}(\vx)$,
because the normal cone is empty, i.e., $N_{\feasSet}(\vx)=\emptyset\text{ for }\vx\notin\feasSet$.
Under standard constraint qualifications, the normal cone admits a representation in terms of dual multipliers
\vspace{-2mm}
\begin{equation}\label{eq:normal_cone_in_dual_form}
N_{\feasSet}(\vx) = \{\matA^\transpose\vy + \matG^\transpose\vz \,\vert\, \vy\in\R^m, \vz\in\R^p_{+}, \vz^\transpose(\matG\vx - \vecH)=\vzero \},  
\end{equation}
from which the~\gls{kkt} conditions follow, with $\vz\ge\vzero$ and $\vz^\transpose(\matG\vx-\vecH)=\vzero$ corresponding to dual feasibility and complementary slackness.

In robotics applications, primal infeasibility commonly arises from over-constrained kinematic tasks or conflicting contact conditions.
In differentiable~\gls{qp} layers, primal feasibility cannot be enforced during training, since intermediate iterates produced by gradient-based updates are generally infeasible.

In the following subsection, we distinguish between violations of dual feasibility at the multiplier level and true dual infeasibility in the optimization-theoretic sense.
\vspace{-2mm}
\subsection{Dual infeasibility}
Dual feasibility concerns the existence of Lagrange multipliers $(\vy,\vz)$ compatible with the normal-cone representation of~\Cref{eq:kkt_inclusion_intro}. It requires non-negativity of the inequality multipliers $\vz \ge \vzero$, defining the dual set $\dualSet = \{(\vy,\vz)\in\R^m\times\R^p \mid \vz \ge \vzero\}$. A pair $(\vy,\vz)$ is dual infeasible if $(\vy,\vz)\notin\dualSet$, and we quantify the violation by the Euclidean distance to $\dualSet$,
\vspace{-2mm}
\begin{equation}\label{eq:dual_infeas_dist}
\mathrm{dist}((\vy,\vz), \dualSet) = \|\vz - \proj{\R^p_{+}}(\vz)\|,
\vspace{-2mm}
\end{equation}

where $\proj{\R^p_{+}}(\cdot)$ denotes the Euclidean projection onto the nonnegative orthant. When $\vz\not\ge\vzero$, the vector $\matA^\transpose\vy+\matG^\transpose\vz$ no longer lies in $N_{\feasSet}(\vx)$, so the stationarity inclusion $\vzero \in \matQ\vx + \vecC + N_{\feasSet}(\vx)$ fails to characterize a valid~\gls{kkt} point, even when $\vx$ is primal feasible. Dual feasibility is therefore necessary for the~\gls{kkt} system to be well defined.

Under standard regularity assumptions such as Slater's condition, dual infeasibility manifests as primal unboundedness, meaning the objective decreases without bound along a feasible direction, $\inf_{\vx\in\feasSet} \tfrac{1}{2}\,\vx^\transpose \matQ\,\vx + \vecC^\transpose \vx = -\infty$. When $\matQ\succeq \vzero$, this requires a nonzero $\Delta\vx$ in the recession cone of $\feasSet$ along which the quadratic term vanishes, i.e. $\matQ\Delta\vx=\vzero$, $\matA\Delta\vx=\vzero$, $\matG\Delta\vx\le\vzero$, and $\vecC^\transpose\Delta\vx<0$. Such directions are rare in well-posed robotics problems and typically signal a modeling pathology, such as missing bounds, or a differentiable~\gls{qp} layer whose learned $\matQ$ has drifted from positive semidefiniteness during training. A small regularization term suffices in both cases. Indefinite $\matQ$ is qualitatively different, since the problem leaves the convex setting altogether, and the resulting unboundedness reflects a loss of numerical validity rather than dual infeasibility in any meaningful sense.

\subsection{About certifying infeasibility}
Reliable infeasibility detection is challenging, since residual-based criteria often stagnate in ill-conditioned or degenerate problems. Formal detection relies on Farkas-type certificates~\cite{banjac2019infeasibility}, which prove unsolvability geometrically but are numerically fragile and expensive to compute post hoc. A common alternative is the homogeneous self-dual formulation~\cite{ye1994}, embedding the~\gls{kkt} system into a skew-symmetric system where optimality and infeasibility appear as distinct limit points of a single central path. Neither approach aligns with the \emph{best-effort} solutions required in robotics and control, where approximate feasibility and numerical robustness outweigh exact infeasibility proofs.
\vspace{-1mm}
\section{\textsc{Elastic Odyn}}\label{sec:method}
\subsection{Algorithmic foundations}\label{sec:elastic_qp}
Elastic formulations address infeasibility by relaxing hard constraints and penalizing violations in the objective. A prototypical elastic~\gls{qp} with explicit infeasibility relaxation reads
\begin{equation}\label{eq:elastic_qp}
\begin{aligned}
\min_{\vx,\,\vse,\,\vsi} \quad
& \tfrac{1}{2}\vx^\transpose \matQ \vx + \vecC^\transpose \vx + \tfrac{1}{2\gamma_e}\norm{\vse}^2 + \tfrac{1}{2\gamma_i}\norm{\vsi}^2 \\
\text{subject to} \quad
& \matA \vx - \vecB = \vse, \quad \matG \vx - \vecH \leq \vsi,
\end{aligned}
\end{equation}
where $\vse\in\R^m$ and $\vsi\in\R^p$ are elastic relaxation variables relaxing the equality and inequality constraints, and $\gamme,\gammi\in\R_+$ balance optimality against constraint satisfaction. Closed-form elimination of $\vse$ and $\vsi$ yields the unconstrained problem
\begin{equation}\label{eq:elastic_qp_eliminated}
\min_{\vx} \tfrac{1}{2}\vx^\transpose \matQ \vx + \vecC^\transpose \vx + \tfrac{1}{2\gamma_e}\norm{\matA\vx-\vecB}^2 + \tfrac{1}{2\gamma_i}\norm{(\matG\vx-\vecH)_+}^2,
\end{equation}
with $(\cdot)_+$ the componentwise positive part. Variationally, the two penalties are Moreau--Yosida regularizations of the indicators $\delta_{\{\bzero\}}$ and $\delta_{\R^p_-}$ composed with the constraint maps,
\begin{equation}\label{eq:variational_elastic_qp_eliminated}
\min_{\vx} \tfrac{1}{2}\vx^\transpose \matQ \vx + \vecC^\transpose \vx + \delta_{\{\bzero\}}^{\gamma_e}(\matA\vx-\vecB) + \delta_{\R^p_-}^{\gamma_i}(\matG\vx-\vecH),
\end{equation}
where $\delta_{\convexSet}^\gamma(\vx)\coloneq\min_{\vx'} \delta_{\convexSet}(\vx') + \tfrac{1}{2\gamma}\|\vx' -\vx\|^2 = \tfrac{1}{2\gamma}\mathrm{dist}^2(\vx,\convexSet)$ is the Moreau envelope of $\delta_{\convexSet}$~\cite{parikh2013proximal}. Its gradient $\nabla\delta_{\convexSet}^\gamma(\vx) = \gamma^{-1}(\vx - \proj{\convexSet}(\vx))$ is the Yosida approximation of the normal cone $N_{\convexSet}$, so the elastic objective is $C^1$ with semismooth gradient, and the first-order optimality condition becomes the semismooth stationarity equation
\begin{equation}\label{eq:elastic_kkt_inclusion}
\vzero = \matQ\vx + \vecC + \tfrac{1}{\gamma_e}\matA^\transpose(\matA\vx-\vecB) + \tfrac{1}{\gamma_i}\matG^\transpose(\matG\vx-\vecH)_+,
\end{equation}
a single-valued surrogate of the original~\gls{kkt} inclusion. Whereas $\ell_1$ penalties enforce constraints exactly at finite parameters\reb{, they lead to} nonsmooth, set-valued optimality conditions\reb{. In contrast,} the squared $\ell_2$ penalties \reb{adopted here} preserve smoothness and yield well-defined proximal~\gls{kkt} systems, \reb{making them} particularly suited to differentiable optimization layers and implicit Newton-type solvers.

\subsection{An elastic path-following NIPM for QP}\label{sec:odyn}

\textsc{Elastic Odyn} is a primal–dual~\gls{qp} solver based on a non-interior-point formulation that handles both feasible and infeasible~\glspl{qp}, supports warm starting, and remains robust to degeneracy. It combines all-shifted~\gls{ncp} functions with proximal primal–dual regularization, yielding a smooth, well-conditioned~\gls{kkt} system under Slater's condition alone:
\vspace{-2mm}
\begin{equation}
\label{eq:elastic_odyn_qp}
\begin{aligned}
\min_{\vx,\, \vs,\, \vse,\, \vsi,\, \vxi} \quad
& \mathbf{f}(\vx) - \mu \sum^p_{j=1}\log(\xi_j) + \tfrac{1}{2\gamme}\|\vse\|^2 + \tfrac{1}{2\gammi}\|\vsi\|^2 \\
\text{subject to} \quad
& \matA \vx - \vecB = \vse, \quad \matG \vx + \vs - \vecH = \vsi, \quad \vs = \vxi,
\end{aligned}
\end{equation}
with $\mathbf{f}(\vx)\coloneq\tfrac{1}{2}\vx^\transpose \matQ \vx + \vecC^\transpose \vx$, inequality slack $\vs\in\R^p$, and consensus variable $\vxi\in\R^p$. The barrier acts on $\vxi$ rather than $\vs$, allowing iterates to leave the interior of $\{\matG\vx \le \vecH\}$ while preserving a well-defined primal–dual structure. We refer to this as a \emph{penalty–barrier formulation}.

The associated Lagrangian augments the consensus constraint quadratically,
\vspace{-2mm}
\begin{equation}
\label{eq:elastic_odyn_lagrangian}
\begin{aligned}
\calP_{\mathcal{E}}(\Ut; \mu, \boldsymbol\gamma)
&= \mathbf{f}(\vx) - \mu \sum_{j=1}^{p} \log(\xi_j) + \tfrac{1}{2\gamma_e}\|\vse\|^2 + \tfrac{1}{2\gamma_i}\|\vsi\|^2 \\
&\quad + \tfrac{1}{2}\|\vs - \vxi\|^2 - \vw^\transpose (\vs - \vxi) \\
&\quad + \vy^\transpose(\matA\vx - \vecB - \vse) + \vz^\transpose(\matG\vx + \vs - \vecH - \vsi),
\end{aligned}
\end{equation}
where $\vy\in\R^m$, $\vz\in\R^p$, $\vw\in\R^p$ are multipliers for the equality, inequality, and consensus constraints, $\mu\in\R_+$ is the barrier parameter, and $\Ut \coloneq (\vx, \vs, \vxi, \vw, \vy, \vz, \vse, \vsi)$.

For robustness under degeneracy, we add proximal penalties on primal--dual variables, yielding the perturbed Lagrangian
\begin{equation}
\label{eq:all_shifted_odyn_langrangian}
\begin{aligned}
\tilde{\calP}_{\mathcal{E}}(&\Ut; \mu, \boldsymbol\gamma, \boldsymbol{\rho}, \Ut_E)
= \calP_{\mathcal{E}}(\Ut; \mu, \boldsymbol\gamma)
+ \tfrac{\rhod}{2}\|\vx - \xE\|^2 \\
&\quad - \tfrac{\rhoe}{2}\|\vy - \yE\|^2 - \tfrac{\rhoi}{2}\|\vz - \zE\|^2 + \tfrac{\rhon}{2}\|\vs- \sE\|^2\\
&\quad-\tfrac{\rhon}{2}\|\vw - \wE\|^2 - \tfrac{\rhon}{2}(\vs - \sE)^\transpose(\vw - \wE),
\end{aligned}
\end{equation}
with reference estimates $\Ut_E\coloneq(\xE, \sE, \wE, \yE, \zE)$ and proximal weights $\boldsymbol{\rho} \coloneq (\rhod, \rhoe, \rhoi, \rhon)\in\R^4_+$. The opposing signs on dual variables reflect the saddle structure of the primal–dual formulation. \reb{Its first order optimality conditions are similar to \textsc{Odyn}'s in~\cite{rojas_odyn_2026} with the additions of~\eqref{eq:all_shifted_elastic_KKT_conditions} and form the basis for \textsc{Elastic Odyn}'s Newton system.
\begin{equation}\label{eq:all_shifted_elastic_KKT_conditions}
\begin{aligned}
\nabla_{\vy}\tilde{\calP}_{\mathcal{E}} &= \matA\vx - \vecB - \vse - \rhoe(\vy - \yE), \\
\nabla_{\vz}\tilde{\calP}_{\mathcal{E}} &= \matG\vx + \vs - \vecH - \vsi - \rhoi(\vz - \zE), \\
\nabla_{\vse}\tilde{\calP}_{\mathcal{E}} &= \vy - \gamme^{-1}\vse, \quad
\nabla_{\vsi}\tilde{\calP}_{\mathcal{E}} = \vz - \gammi^{-1}\vsi.
\end{aligned}
\vspace{-5mm}
\end{equation}}
\subsection{Elastic Newton system}
Following \textsc{Odyn}'s reduction strategy~\cite{rojas_odyn_2026}, we first eliminate the consensus variable $\vxi$ in closed form by substituting $\nabla_{\vw}\tilde{\calP}_{\mathcal{E}}=\vzero$ into $\nabla_{\vs}\tilde{\calP}_{\mathcal{E}}=\vzero$, simplifying the corresponding~\gls{kkt} residuals. The elastic relaxation variables are likewise eliminated via the identities $\vse\equiv\gamma_e \vy$ and $\vsi\equiv\gamma_i \vz$ in~\Cref{eq:all_shifted_elastic_KKT_conditions}. This twofold elimination reduces the system to the hard-constrained dimensions and removes the need for an explicit non-negativity constraint on $\vsi$, which inherits non-negativity from the inequality multiplier $\vz$.

These reductions yield the following set of reduced elastic~\gls{kkt} residuals:
\begin{equation}\label{eq:elastic_KKT_residual}
\tilde{\vr}(\U;\boldsymbol{\Pi}) = 
\begin{bmatrix}
\tilde{\vr}_d\\
\tilde{\vr}_e\\
\tilde{\vr}_i\\
\tilde{\vr}_g
\end{bmatrix}
=
\begin{bmatrix}
\matQ \vx + \vecC + \matA^\transpose \vy + \matG^\transpose \vz + \rhod(\vx - \xE)\\
\matA \vx - \vecB - \gamme \vy - \rhoe(\vy - \yE)\\
\matG \vx + \vs - \vecH - \gammi \vz - \rhoi(\vz - \zE)\\
\vncpFun{}(\vs,\vz;\mu) + \rhon(\vs - \sE) + \rhon(\vz - \zE)
\end{bmatrix}
\end{equation}
where $\vncpFun{} : \R^{p} \times \R^{p} \to \R^{p}$ is an~\gls{ncp} function, $\U \coloneq \left(\vx, \vs, \vy, \vz\right)\in\R^{n+m+2p}$ is the reduced set of primal and dual variables, and $\boldsymbol{\Pi} \coloneq \left(\mu, \boldsymbol \gamma, \boldsymbol{\rho}, \U_E\right)$ denotes the set of parameters of the elastic penalty--barrier formulation.
\Cref{eq:elastic_KKT_residual} is then solved by applying Newton's method:
\begin{equation}\label{eq:elastic_KKT_system}
\begin{bmatrix}
\Qtil      & \matA^\transpose                 & \matG^\transpose &  \\
\matA      & -\left(\rhoe + \gamme\right) \Id &           &  \\
\matG      &                                  & -\left(\rhoi + \gammi\right) \Id & \Id \\
           &                                  & \pertncpJac{\vz} & \pertncpJac{\vs}
\end{bmatrix}
\begin{bmatrix}
\dx\\
\dy\\
\dz\\
\ds
\end{bmatrix}
=
-
\begin{bmatrix}
\tilde{\vr}_d\\
\tilde{\vr}_e\\
\tilde{\vr}_i\\
\tilde{\vr}_g
\end{bmatrix},
\end{equation}
where $\pertncpJac{\vz} = \ncpJac{\vz} + \rho_n\Id$ and $\pertncpJac{\vs} = \ncpJac{\vs} + \rho_n\Id$ are diagonal matrices in which $\ncpJac{\vz,\vs}\in\R^{p\times p}$ denotes the~\gls{ncp} Jacobians.

From~\Cref{eq:elastic_KKT_residual}, the elastic formulation penalizes equality and inequality violations through scaled Lagrange multipliers, whereas in~\Cref{eq:elastic_KKT_system} these penalties act as adaptive regularization terms whose magnitudes are governed by the current~\gls{kkt} residuals. \textsc{Elastic Odyn} factorizes this system and accepts the resulting search directions via a non-monotone line search (see Section~IV.D in~\cite{rojas_odyn_2026}). 
\reb{Note that~\Cref{eq:elastic_KKT_system} has the same dimensions as the hard constrained system, requiring one $n\times n$ Cholesky factorization per iteration. The elastic penalties affect only the diagonal terms $(\rhoe+\gamme)\Id$ and $(\rhoi+\gammi)\Id$. Thus, the per iteration cost matches that of \textsc{Odyn} up to $O(m+p)$ diagonal operations, and does not alter the scalability properties established in~\cite{rojas_odyn_2026}.}

\subsubsection*{Barrier, estimates, and elastic penalty updates}
\reb{
The barrier parameter smooths the~\gls{ncp} functions via the centering procedure of~\cite{kanzow1996}. At the end of each iteration, the proximal estimates interpolate between $\U$ and $\U^+$, transitioning from conservative gradient-descent-like updates to Newton-like steps as sufficient progress is achieved~\cite{rojas_odyn_2026}, similarly to Levenberg--Marquardt schemes. Interpolation parameters and elastic penalties $\gamme,\gammi$ tighten when feasibility progress is sufficient and remain unchanged otherwise.
}

\subsection{Stopping Criteria}

We combine absolute and relative tolerances to certify convergence in both feasible and infeasible regimes. Unlike standard~\glspl{qp} solvers, \textsc{Elastic Odyn}'s stopping criteria remain well defined under infeasibility and guarantee convergence to the \emph{closest feasible} problem when exact feasibility cannot be attained.

The primal residuals $(\norm{\vr_e}_\infty, \norm{\vr_i}_\infty)$ and dual residual $\norm{\vr_d}_\infty$ are scaled to reflect the elastic relaxation. Primal relative tolerances incorporate $\gamma_e\|\vy\|_\infty$ and $\gamma_i\|\vz\|_\infty$, matching the relaxed residuals via the identities $\vse = \gamma_e\vy$ and $\vsi = \gamma_i\vz$ in~\Cref{eq:all_shifted_elastic_KKT_conditions}. The dual relative tolerance is symmetrically augmented by $\gamma_e^{-1}\|\vse\|_\infty$ and $\gamma_i^{-1}\|\vsi\|_\infty$, ensuring that feasibility and optimality are evaluated consistently with the relaxed problem throughout the iteration.

To certify convergence to the closest-feasible~\gls{qp}, we further enforce a least-violation optimality condition,
\begin{equation}
\begin{aligned}
\norm{\vr_c}_\infty 
\leq \epsilon_a + \epsilon_r \max\!\Big(
\norm{\matA^\transpose \matA \vx + \matG^\transpose (\matG \vx + \vs)}_\infty,\\
\norm{\matA^\transpose \vecB}_\infty,\;\norm{\matG^\transpose \vecH}_\infty,\;
\norm{\matA^\transpose \vy + \matG^\transpose \vz}_\infty
\Big),
\end{aligned}
\end{equation}
where $\vr_c$ is the gradient of the squared constraint violation,
\begin{equation}
\label{eq:gradient_residual}
\vr_c = \nabla_{\vx}\!\Big(\tfrac12\norm{\matA\vx - \vecB}^2 + \tfrac12\norm{\matG \vx + \vs - \vecH}^2\Big)
\end{equation}
This certifies optimality in the \emph{least-violation} squared $\ell_2$ sense. Combined with the elastic primal and dual scalings, these criteria guarantee convergence to the optimally shifted problem while preserving a meaningful primal–dual solution under infeasibility.
\vspace{-3mm}

\subsection{Warm-started refining stage}\label{sec:refining_stage}
The elastic formulation robustly recovers a primal solution for the closest-feasible~\gls{qp}, but its dual variables are shaped by the elastic stationarity in~\Cref{eq:all_shifted_elastic_KKT_conditions} and encode penalized violations rather than true sensitivities of the shifted feasible constraints. These elastic duals therefore lack physical interpretability, making them unsuitable as Lagrange multipliers in contact-physics settings, where multipliers represent quantities like contact forces. We address this with a lightweight refinement stage that warm-starts from the elastic primal, shifts the constraints with the optimal elastic relaxation variables, and re-solves the hard-constrained problem. Converging in only a few iterations (see~\Cref{sec:qp_problems}), it yields a well-conditioned primal–dual pair with physically meaningful multipliers and sensitivities consistent with the closest-feasible problem.
\vspace{-4mm}
\subsection{Differentiable optimization under infeasibility}\label{sec:diff_qp}

\textsc{Elastic OdynLayer} extends differentiable optimization to \emph{infeasible}~\glspl{qp}, allowing gradients to propagate when constraints conflict and exposing their violations as learning signals. Unlike feasibility-assuming layers~\cite{amos2017optnet,cvxpylayers2019}, it differentiates through the parameterized elastic~\gls{qp}
\begin{equation}
\label{eq:diff_elastic_qp}
\begin{aligned}
\min_{\vx,\vse,\vsi}\quad
& \tfrac12 \vx^\transpose \matQ(\vtheta)\vx
+ \vecC(\vtheta)^\transpose\vx
+ \tfrac{1}{2\gamma_e}\|\vse\|^2
+ \tfrac{1}{2\gamma_i}\|\vsi\|^2 \\
\text{subject to}\quad
& \matA(\vtheta)\vx-\vecB(\vtheta)=\vse,\quad
\matG(\vtheta)\vx-\vecH(\vtheta)\leq\vsi,
\end{aligned}
\end{equation}
where $\vtheta\in\R^r$ collects the learnable parameters. In contrast to the predict-and-optimize works surveyed in~\cite{kotary2021survey}, which parametrize the objective and hold the constraints fixed, all constraint data \reb{$\matA, \vecB, \matG, \vecH$} are learnable here.\reb{ To the best of our knowledge, QPLayer~\cite{bambade-2024qplayer} is the only differntiable QP solver that hanldes infeasibility in the literature.}
\paragraph{Forward pass}
\reb{
\textsc{Elastic Odyn} solves~\Cref{eq:diff_elastic_qp}, or its
smoothed KKT system for a positive target $\mu$, returning
$\U^\star$ and the violation measures $\vse^\star=\matA\vx^\star-\vecB, \quad \vsi^\star=(\matG\vx^\star-\vecH)_+$, which may enter the training loss directly.

\paragraph{Backward pass}
At regular solutions, we differentiate the elastic KKT residual
$\vr$ with final penalties and smoothing fixed and algorithmic
proximal terms removed:
\begin{equation}
\label{eq:odynlayer_adjoint}
\left[\frac{\partial\vr}{\partial\U^\star}\right]^\transpose
\boldsymbol{\lambda}
=-\frac{\partial\ell}{\partial\U^\star},
\end{equation}
yielding the implicit gradient
$\boldsymbol{\lambda}^\transpose\partial\vr/\partial\vtheta$.
Bounded complementarity weights yield a symmetric system of
dimension $n+m+p$, which is solved by a single factorization and residual
correction without Hessian shifts. Singular systems trigger
a warned least-squares selection in scaled reduced coordinates.

\paragraph{Infeasibility-aware losses}
For $\vge\coloneq\partial\ell/\partial\vse^\star$ and
$\vgi\coloneq\partial\ell/\partial\vsi^\star$, the chain rule
through the positive-part operator and $\mathbf m_{g_i} := \mathbf{1}_{\vsi^\star>0}\odot\vgi$ gives
\begin{equation}
\label{eq:slack_grads}
\begin{aligned}
\frac{\partial\ell}{\partial\vy^\star}
&\mathrel{+}=\gamma_e\vge,\quad \frac{\partial\ell}{\partial\vz^\star} \mathrel{+}=\gamma_i\mathbf m_{g_i}, \quad
\frac{\partial\ell}{\partial\vs^\star} \mathrel{-}=\mathbf m_{g_i}.
\end{aligned}
\end{equation}
These sensitivities enter the same adjoint, enabling losses
that penalize inconsistency without an additional backward procedure.}

\begin{figure*}[ht]
    \centering
        \begin{minipage}[t]{0.33\textwidth}
            \vspace{0pt}
            \centering
            \includegraphics[height=0.12\textheight, trim=0 0 0 0, clip]{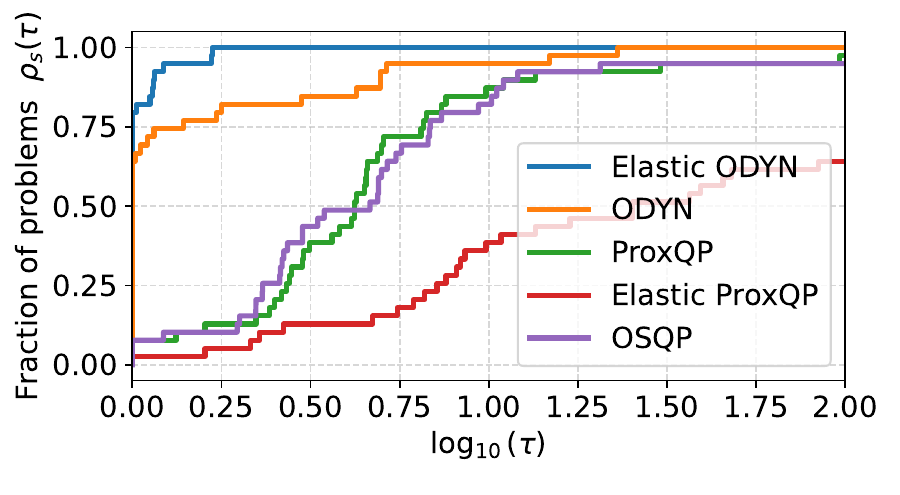}
            \vspace{-2mm}
            \par\footnotesize\textbf{(a)} Perturbation $\Delta=0.1$
        \end{minipage}%
        \hfill
        \begin{minipage}[t]{0.33\textwidth}
            \vspace{0pt}
            \centering
            \includegraphics[height=0.12\textheight,, trim=0 0 0 0, clip]{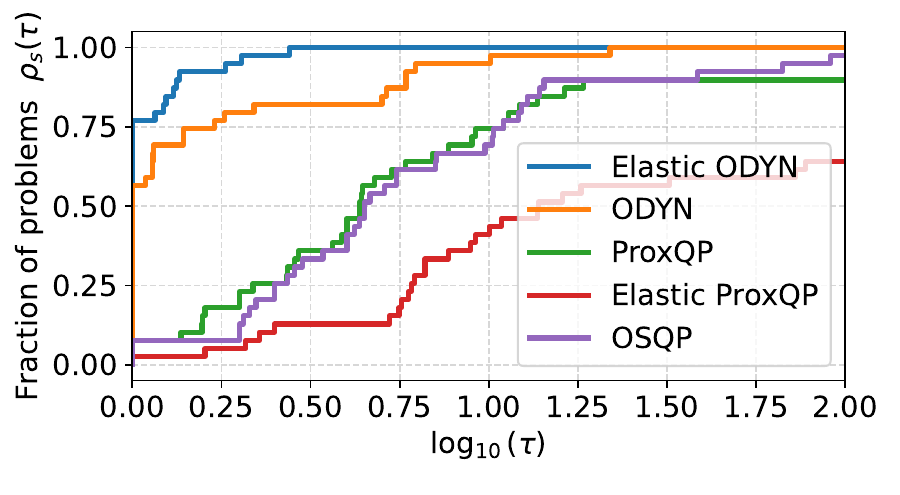}
            \vspace{-2mm}
            \par\footnotesize\textbf{(b)} Perturbation $\Delta=0.01$
        \end{minipage}%
        \begin{minipage}[t]{0.33\textwidth}
            \vspace{0pt}
            \centering
            \includegraphics[height=0.12\textheight, trim=0 0 0 0, clip]{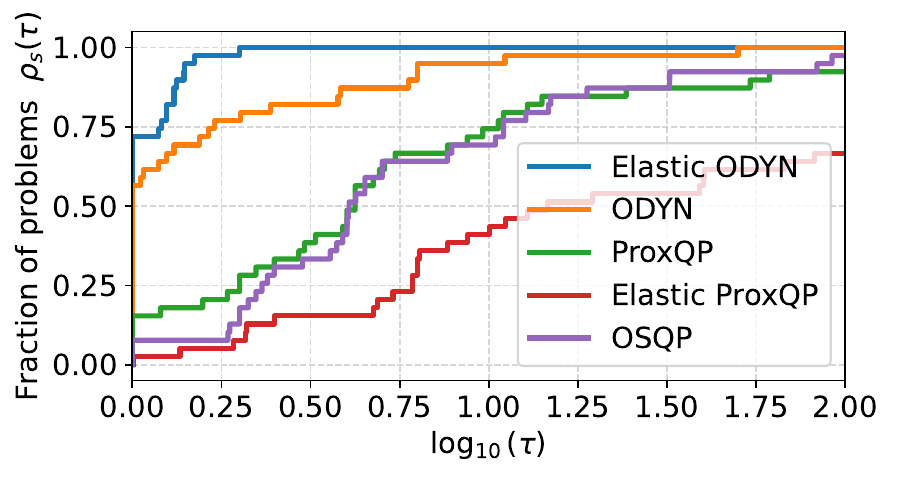}
            \vspace{-2mm}
            \par\footnotesize\textbf{(c)} Perturbation $\Delta=0.001$
        \end{minipage}%
        \hfill
    \caption{Warm-start performance on the perturbed Maros--Mészáros test set under high-accuracy settings.
    We compare \textsc{Elastic Odyn}, \textsc{Odyn}, \textsc{ProxQP}, \textsc{Elastic ProxQP}, and \textsc{OSQP}. \textsc{Elastic Odyn} consistently outperforms all competing solvers, both hard-constrained and elastic variants.}
    \label{fig:warmstart_results}
    \vspace{-5mm}
\end{figure*}
\vspace{-4mm}
\section{Results}\label{sec:results}
This section evaluates \textsc{Elastic Odyn} on Maros–Mészáros benchmarks, infeasible problems, contact mechanics, learning dynamics, and trajectory optimization.
\reb{All experiments are simulated. Hardware results and solver-specific ablations are reported in \textsc{Odyn}~\cite{rojas_odyn_2026}, whose computational budget and \gls{nipm} machinery carries over unchanged (\Cref{sec:method}). \textsc{Elastic Odyn}-specific hardware evaluation is left for future work.}

\subsection{The Maros--Mészáros test set of hard QPs}\label{sec:maros_meczaros}
We benchmark \textsc{Elastic Odyn} against state-of-the-art solvers with $\ell_2$ elastic formulations, namely \textsc{Elastic ProxQP}, which naturally converges to the closest-feasible~\gls{qp} under contradicting constraints, and \textsc{Gurobi}, which employs a two-stage restoration process. We compare convergence and warm-starting performance.
{
\vspace{-3mm}
\setlength{\tabcolsep}{4pt}
\begin{table}[ht]
\caption{Performance for hard and elastic solver variants.}
\vspace{-3mm}
\label{tab:failure_rate}
\centering
\begin{tabular}{@{}l c|c c|c c|c@{}}
\toprule
\textbf{Failure Rates [\%]}
& \multicolumn{2}{c}{\textbf{\textsc{Odyn}}}
& \multicolumn{2}{c}{\textbf{\textsc{ProxQP}}}
& \multicolumn{2}{c}{\textbf{\textsc{Gurobi}}} \\
& Hard & Elastic & Hard & Elastic & Hard & Elastic \\
\textbf{Medium} $[10^{-6}]$
& \textbf{15.5} & \textbf{3.9}
& 40.8 & 9.7
& \textbf{15.5} & 10.7 \\
\textbf{High} $[10^{-9}]$
& \textbf{16.5} & 15.5
& 44.7 & 10.7
& \textbf{16.5} & \textbf{7.8} \\
\midrule
\textbf{Median per iter. [ms]}
& \multicolumn{2}{c}{$\mathbf{0.48}$}
& \multicolumn{2}{c}{$0.51$}
& \multicolumn{2}{c}{$1.88$} \\
\textbf{Mean per iter. $(\pm \sigma)$ [ms]}
& \multicolumn{2}{c}{$\mathbf{4.5\!\pm\!14.0}$}
& \multicolumn{2}{c}{$30.2\!\pm\!80.5$}
& \multicolumn{2}{c}{$41.9\!\pm\!49.0$} \\
\bottomrule
\vspace{-6mm}
\end{tabular}
\end{table}
}

\subsubsection{Convergence}\label{sec:maros_meczaros_failure}
\reb{\Cref{tab:failure_rate} reports failure rates at medium and high accuracy for hard and elastic formulations across solvers supporting elastic constraints. \textsc{Elastic Odyn} achieves the lowest failure rate at medium accuracy, demonstrating robustness in regimes typical of robotics and AI. At high accuracy, \textsc{Gurobi} outperforms \textsc{ProxQP}, while \textsc{Elastic Odyn}'s convergence deficit is largely attributable to its lack of iterative refinement and preconditioning, both used by its competitors. On the 81 problems solved by all methods, \textsc{Elastic Odyn} achieves the fastest per-iteration runtime.}

\subsubsection{Warm-starting performance}\label{sec:maros_meczaros_warmstartign}
We evaluate warm-start performance of \textsc{Elastic Odyn} against \textsc{Odyn}, \textsc{OSQP}, \textsc{ProxQP}, and \textsc{Elastic ProxQP} on Maros--Mészáros problems at high accuracy. Interior-point solvers such as \textsc{Gurobi} are excluded for lacking warm-start support, and \textsc{OSQP} is restricted to hard constraints. Following the protocol of~\cite{rojas_odyn_2026}, each problem is solved from a cold start, then re-solved on perturbed instances of magnitude $\delta\in\{10^{-3},10^{-2},10^{-1}\}$ that preserve sparsity and symmetry, warm-started from the original primal--dual solution. Performance is measured via the warm-to-cold iteration ratio. \Cref{fig:warmstart_results} shows that \textsc{Elastic Odyn} achieves the lowest warm-start iteration counts among all compared methods.
\vspace{-2mm}
\begin{table}[h]
\centering
\caption{Convergence comparison of \textsc{Odyn}, \textsc{ProxQP} and \textsc{Gurobi}}
\label{tab:odyn_vs_proxqp}
\vspace{-3mm}
\begin{tabular}{@{}c*{3}{c}@{}}
\toprule
\textbf{Problem}  & \textbf{\textsc{Odyn}} & \textbf{\textsc{ProxQP}} & \textbf{\textsc{Gurobi}}\\
tolerance: $10^{-9}$ & iterations + refinement & iterations & iterations \\
\midrule
$\feasSet_1$    & \textbf{28 + 1}  & \textit{MaxIter} & \textit{MaxIter}\\
$\feasSet_2$    & \textbf{98 + 10} & \textit{MaxIter} & \textit{MaxIter}\\
$\feasSet_3$    & \textbf{102 + 1} & \textit{MaxIter} & \textit{MaxIter}\\
$\feasSet_4$    & 87 + 2 & \textit{MaxIter} & \textbf{56}\\
$\feasSet_5$    & 92 + 4 & \textit{MaxIter} & \textbf{22}\\
\bottomrule
\end{tabular}
\vspace{-4mm}
\end{table}

\subsection{Ill-conditioned and degenerate infeasible QPs}\label{sec:qp_problems}
We evaluate \textsc{Elastic Odyn} on a set of deliberately infeasible and ill-conditioned~\glspl{qp} designed to stress numerical robustness, conditioning, and convergence. We compare against \textsc{ProxQP} and \textsc{Gurobi}, both of which feature $\ell_2$ restoration and resolution stages. Each test isolates a specific failure mode, namely inconsistent constraints, poor scaling, degeneracy, and mixed equality and inequality structures. \Cref{tab:odyn_vs_proxqp} reports convergence and solution quality, highlighting the robustness of \textsc{Elastic Odyn} under infeasibility.
The constraint sets are,
\vspace{-1mm}
\begin{equation*}
\begin{aligned}
&\feasSet_1 \coloneq \left\{(x_1,x_2)\in\R^2 \;\middle|\; x_1 = 1,\; x_1 = -2,\; x_2 = 1,\; x_2 = 10^3 \right\}\\
&\feasSet_2 \coloneq \left\{(x_1,x_2)\in\R^2 \;\middle|\; x_1 \le 2,\; x_1 \ge 1{,}200{,}000\right\}\\
&\feasSet_3 \coloneq \left\{(x_1,x_2)\in\R^2 \;\middle|\; \begin{aligned} & x_1 = 3,\; x_1 = -2,\; x_2 = 1,\\ & x_2 = 200,\; x_1 \le 2,\; x_1 \ge 10\end{aligned}\right\}\\
&\feasSet_4 \coloneq \left\{(x_1,x_2)\in\R^2 \;\middle|\; x_1 = 1,\; x_1 = 1,\; x_1 \ge 2,\; x_1 \ge 32 \right\}\\
&\feasSet_5 \coloneq \left\{ x_1 + x_2 \ge 2.5,\; (x_1, x_2) \in [0, 1]\right\},
\end{aligned}
\vspace{-0.5mm}
\end{equation*}
where $\feasSet_2$ exhibits significant scaling mismatches, $\feasSet_4$ contains linearly dependent constraints, and $\feasSet_5$ pairs an infeasible linear program with the objective $\min_{x_1,x_2}(x_1 + x_2)$, isolating infeasibility arising solely from conflicting bounds under a linear cost.
To verify that \textsc{Elastic Odyn} returns the genuine closest-feasible solution, we compare its primal solutions against the analytical least-squares solution of each $\feasSet_i$. The mean squared errors range from $10^{-4}$ on the severely scaled $\feasSet_2$ down to near machine precision ($10^{-20}$) on $\feasSet_5$, confirming that~\Cref{eq:gradient_residual} certifies optimality in the least-violation $\ell_2$ sense.
\vspace{-5mm}
\begin{figure}[h]
  \centering
  \begin{minipage}[t]{1.0\columnwidth}
    \vspace{0pt}\centering
    \includegraphics[height=0.07\textheight,trim=40 15 0 0,clip]{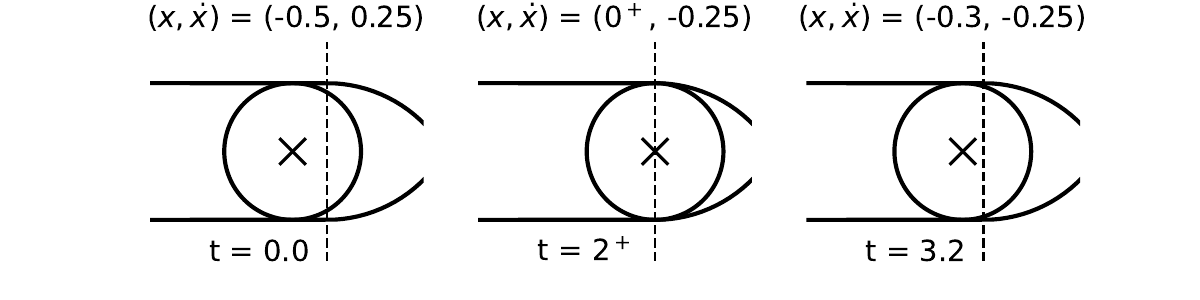}
  \end{minipage}\vfill
  \begin{minipage}[t]{0.49\columnwidth}
    \vspace{1pt}\centering
    \includegraphics[height=0.14\textheight,trim=15 0 0 10,clip]{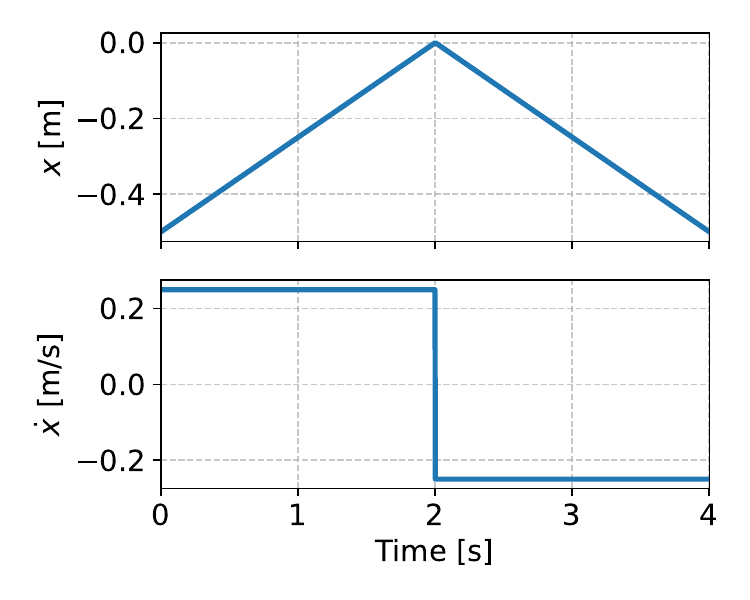}
  \end{minipage}\hfill
  \begin{minipage}[t]{0.509\columnwidth}
    \vspace{0pt}\centering
    \includegraphics[height=0.14\textheight,trim=10 0 15 10,clip]{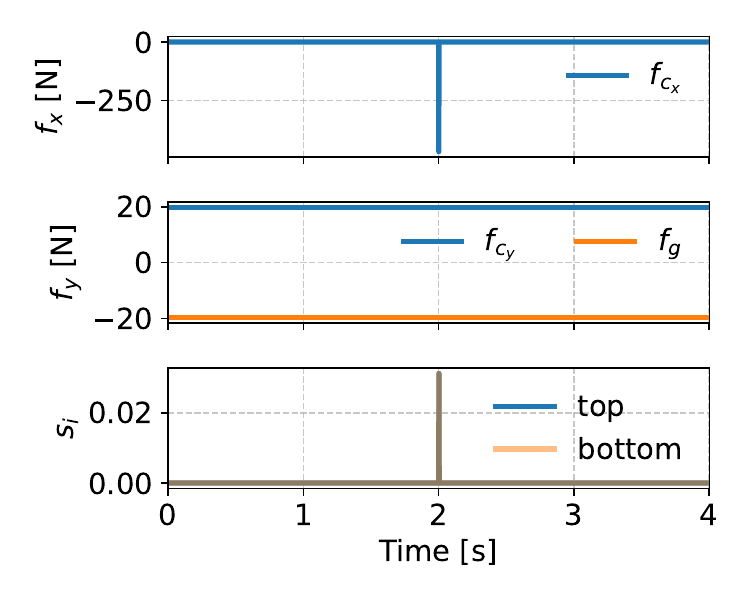}
  \end{minipage}
  \vspace{-5mm}
  \caption{\textbf{Frictionless disc-in-tube simulation}. The disc moves with $\dot x = 0.25$ and reaches a constriction at $x = 0$, where the contact normals are linearly dependent and cannot generate a net force along $x$. \textsc{Elastic Odyn} resolves this elastic impact through a controlled violation $\vsi$, permitting slight penetration that restores force generation along the $x$-direction.}
  \vspace{-4mm}
  \label{fig:contact_sim_results}
\end{figure}

\subsection{Contact mechanics}\label{sec:contact_mechanics}
Contact mechanics provide a canonical setting where infeasibility arises frequently. At the acceleration level, sustained contact and impulsive impact are described by the same~\gls{qp},
\begin{equation}
\begin{aligned}\label{eq:gauss_qp}
\min_{\mathbf p,\,\vsi}\;&
\frac12\|\mathbf p-\mathbf p_{0}\|_{\mathbf M}^2 + \frac{1}{2\gammi}\|\vsi\|^2\\
\text{subject to}\;&
\mathbf J \mathbf p + \mathbf b + \vsi \ge \vzero,
\end{aligned}
\end{equation}
differing only in the temporal interpretation of the constraint geometry. The variables specialize to $(\mathbf p, \mathbf p_0, \vecB) = (\mathbf a, \mathbf a_\text{free}, \dot{\mathbf J}\mathbf v)$ for sustained contact, with generalized acceleration $\mathbf a\in\R^n$, generalized velocity $\mathbf v\in\R^n$, unconstrained acceleration $\mathbf a_\text{free} = \mathbf M^{-1}(\boldsymbol\tau - \mathbf h)$, and contact Jacobian $\mathbf J\in\R^{p\times n}$. For impact, $(\mathbf p, \mathbf p_0, \vecB) = (\mathbf v^+, \mathbf v^-, e\mathbf J\mathbf v^-)$ with pre- and post-impact velocities $\mathbf v^-, \mathbf v^+\in\R^n$ and restitution coefficient $e\in[0,1]$. Sustained contact enforces non-penetration over a finite time step, while impact enforces it instantaneously through impulses. In both cases, infeasibility arises when $\mathbf J$ becomes rank-deficient or geometrically inconsistent, preventing the constraints from generating admissible reaction forces.

The elastic variable $\vsi$ relaxes the non-penetration constraint and penalizes infeasibility in an $\ell_2$ sense. Its complementarity-slackness condition yields the regularized~\gls{lcp}
\begin{equation}
\label{eq:reg_lcp}
\vzero \le \underbrace{(\mathbf{J}\mathbf{M}^{-1}\mathbf{J}^\transpose + \gammi \mathbf{I})}_{\text{reg. Delassus operator}} \vz + \underbrace{\mathbf J\mathbf p_0 + \vecB}_{\text{contact bias}} \perp \vz \ge \vzero,
\end{equation}
where $\vz\in\R^p$ collects the contact forces in sustained-contact models or the collision impulses in impact scenarios. The diagonal regularization $\gammi\mathbf{I}$, together with the proximal-point perturbation on $\vz$, resolves rank deficiencies in the Delassus operator and enables robust contact-implicit~\gls{mpc} and rigid-body simulation.

We illustrate this on a classical example from Featherstone (Fig.~11.3 in~\cite{featherstone}), where a two-dimensional disc is constrained inside a frictionless tube of equal radius. As the disc reaches the constriction at $x \rightarrow 0^{-}$, the upper and lower contact normals become linearly dependent and $\mathbf J$ loses rank (see~\Cref{fig:contact_sim_results}). At the degenerate configuration $x = 0$ the contact model is singular. With the tube contracting and no tangential forces available, no admissible contact forces can prevent further penetration. \textsc{Elastic Odyn} resolves the singularity by admitting a small penetration beyond $x = 0$, which tilts the contact normals away from collinearity and restores Jacobian rank. The impact dynamics are then captured by~\Cref{eq:gauss_qp}.

\begin{figure}[t]
  \centering
  \includegraphics[width=0.494\columnwidth, trim=0 5 15 0, clip]{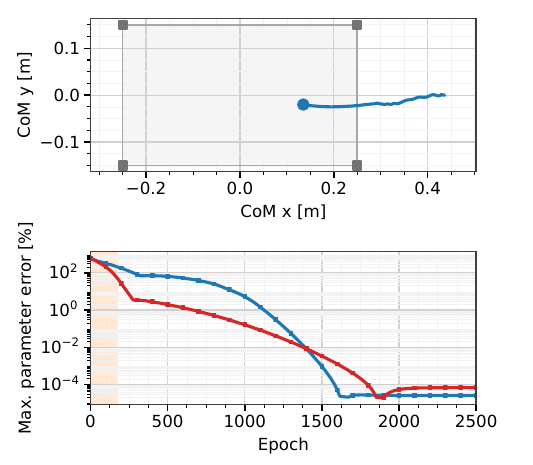}\hfill
  \includegraphics[width=0.497\columnwidth, trim=10 5 5 0, clip]{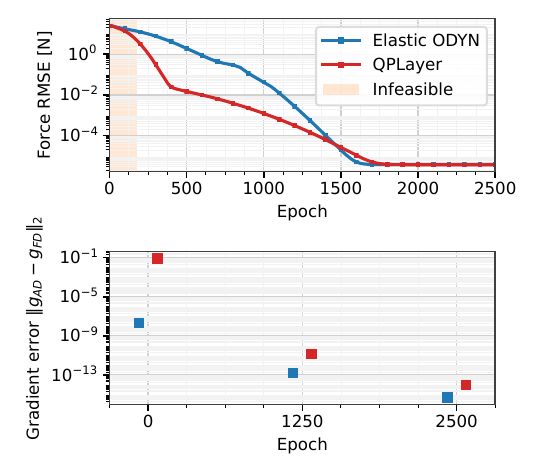}\hfill
  \vspace{-4mm}
  \caption{\reb{Quasi-static load and contact identification with \textsc{Elastic ODYN} and \textsc{QPLayer}. (Top) CoM estimates with the final estimate marked (left), and contact-force RMSE (right). (Bottom) Maximum relative parameter error (left), and worst $\ell_2$ gradient  with finite differences at shared checkpoints (right).}}
  \label{fig:quad_odynlayer}
  \vspace{-7mm}
\end{figure}

\subsection{\textsc{Elastic OdynLayer} for infeasible learning problems}\label{sec:odynlayer}
\reb{
We validate the elastic adjoint against central finite differences on three QP families, each in feasible and infeasible regimes. The degenerate family progressively makes inequality normals nearly linearly dependent, while keeping the Hessian well-conditioned. The ill-conditioned family instead preserves the constraint geometry while increasing the Hessian condition number from $10$ to $10^9$, progressively worsening the numerical conditioning of the QP. A finite difference is retained only if it is stable across four step sizes and all perturbations preserve the elastic penalty, active set, and forward convergence, amounting to about 800 comparisons.
{\vspace{-3mm}
\setlength{\tabcolsep}{2pt}
\begin{table}[ht]
\reb{
\caption{\textsc{Elastic OdynLayer} gradient error relative to finite differences.}
\vspace{-3mm}
\label{tab:gradient_accuracy}
\centering
\small
\resizebox{\columnwidth}{!}{%
\begin{tabular}{@{}l|cc|cc|cc@{}}
\toprule
\textbf{Error}
& \multicolumn{2}{c|}{\textbf{Regular}}
& \multicolumn{2}{c|}{\textbf{Degenerate}}
& \multicolumn{2}{c}{\textbf{Ill-conditioned}} \\
$[10^{-6}]$
& Feas. & Infeas.
& Feas. & Infeas.
& Feas. & Infeas. \\
\midrule
\textbf{Mean} $(\pm \sigma)$
& $0.07\!\pm\!0.67$
& $2.35\!\pm\!8.64$
& $0.04\!\pm\!0.24$
& $3.36\!\pm\!10.95$
& $2.09\!\pm\!8.85$
& $2.20\!\pm\!5.28$ \\
\bottomrule
\end{tabular}%
}
}
\vspace{-3mm}
\end{table}
}

Table~\ref{tab:gradient_accuracy} reports the mean relative gradient error and standard deviation. Across all six regimes, the mean error remains below $3.4\times10^{-6}$, with substantially smaller errors in the regular and degenerate feasible cases.
}

\reb{We evaluate the performance of \textsc{Elastic OdynLayer} by identifying a quadruped's total mass, planar CoM, effective friction coefficient and normal-force limit, $\boldsymbol{\theta}=(m,c_x,c_y,\mu_f,\bar f_z)$, from $36$ quasi-static force observations under a known horizontal tow. Reference forces satisfy exact equilibrium at the true parameters. For each pose $k$, the normalized forces $\mathbf{x}_k=\mathbf{f}_k/F_0\in\mathbb{R}^{12}$ solve
\begin{equation}
\begin{aligned}
\min_{\mathbf{x}_k,\mathbf{s}_{e,k}}\quad&
\tfrac12\|\mathbf{x}_k\|_2^2+
\tfrac{1}{2\gamma_e}\|\mathbf{s}_{e,k}\|_2^2,\\
\text{s.t.}\quad&
\mathbf{A}_k\mathbf{x}_k-\mathbf{b}_k(\boldsymbol{\theta})
=\mathbf{s}_{e,k},\qquad
\mathbf{G}(\boldsymbol{\theta})\mathbf{x}_k
\le\mathbf{h}(\boldsymbol{\theta}).
\end{aligned}
\end{equation}
The six elastic rows encode force and moment balance, while the $24$ hard inequalities impose $0\le f_{ki,z}\le\bar f_z$ and $|f_{ki,x}|,|f_{ki,y}|\le\mu_f f_{ki,z}/\sqrt{2}$, enforcing unilateral support, force limits and conservative friction pyramids. Forces and moments are scaled by $F_0=100\,\mathrm{N}$ and $F_0L$, with $L=0.25\,\mathrm{m}$. Initially, exact equilibrium is infeasible at every pose, with elasticity permiting balance errors while preserving these contact constraints. Both layers minimize
\begin{equation}
\mathcal{L}(\boldsymbol{\theta})=\frac1{36}\sum_{k=1}^{36}
\left[
\frac{\|\mathbf{x}_k(\boldsymbol{\theta})-\hat{\mathbf{x}}_k\|_2^2}{12}
+\frac{\|\mathbf s_{e,k}\|_2^2}{6}
\right],
\end{equation}
where $\hat{\mathbf{x}}_k$ denotes the normalized reference forces and the wrench term directly constrains mass and CoM. \textsc{Elastic Odyn} natively supports \textit{selective elasticity}. We benchmark only against QPLayer, the only alternative supporting infeasibility. QPLayer uses $18$ variables $(\mathbf{x}_k,\mathbf{s}_{e,k})$ with $[\mathbf{A}_k,-\mathbf{I}_6]$ and $[\mathbf{G},\mathbf{0}]$, relaxing balance while preserving hard contact inequalities. Both use $\gamma_e=5\times10^{-7}$, forward tolerances of $10^{-9}$, and $2500$ batched Adam epochs at learning rate $10^{-2}$. Figure~\ref{fig:quad_odynlayer} shows that \textsc{Elastic Odyn} provides substantially more accurate gradients under infeasibility, with comparable convergence and slightly better final accuracy.}

\begin{figure}[tb]
\centering
\ifpublished
\href{https://youtu.be/X9AgWGu5Eug}
{
\includegraphics[width=0.65\linewidth]{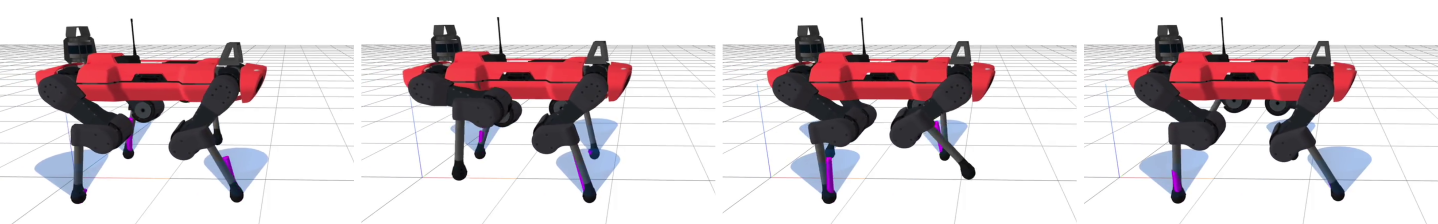}\\
\includegraphics[width=0.405\linewidth, trim=0 0 0 0, clip]{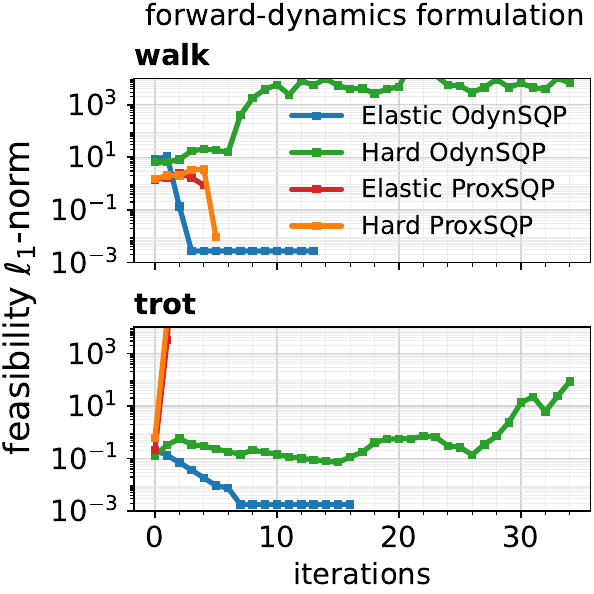}
\includegraphics[width=0.32\linewidth, trim=60 0 0 0, clip]{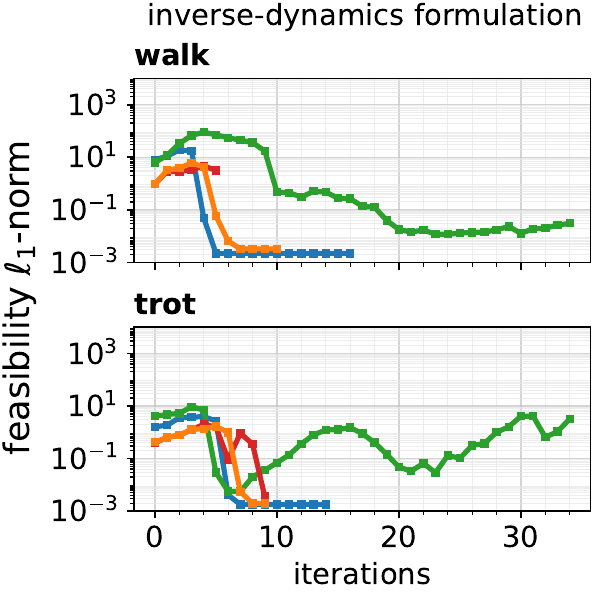}}
\else
\includegraphics[width=0.65\linewidth]{Figures/anymal_snapshots.png}\\
\includegraphics[width=0.405\linewidth, trim=0 0 0 0, clip]{Figures/feas_forward.pdf}
\includegraphics[width=0.32\linewidth, trim=60 0 0 0, clip]{Figures/feas_inverse.pdf}
\fi
\vspace{-3mm}
\caption{Feasibility evolution for elastic and hard formulations in a highly constrained quadrupedal locomotion task.
(Top) Snapshots of the ANYmal walking motion computed using our SQP framework.
(Bottom) Forward and inverse-dynamics optimal control feasibilities.
}
\label{fig:sqp_results}
\vspace{-7mm}
\end{figure}
\vspace{-1mm}
\subsection{\textsc{Elastic OdynSQP} for infeasible optimal control}\label{sec:odyn_sqp}
We extend \textsc{OdynSQP} (Section~V in~\cite{rojas_odyn_2026}) to handle infeasible QP subproblems from inconsistent SQP linearizations and intrinsically infeasible optimal control problems from conflicting nonlinear constraints. Whereas standard SQP cannot compute a step when its QP subproblem is infeasible, \textsc{Elastic OdynSQP} relaxes constraints to maintain solvability.

\textbf{Elastic inner relaxation.} For inconsistent linearizations of otherwise feasible nonlinear problems, the outer loop behaves as standard \textsc{OdynSQP}, while the inner \textsc{Elastic Odyn} resolves infeasible QP subproblems. On a quadrupedal locomotion task with dynamics, contact, torque, and kinematic \textit{hard constraints}, elastic relaxation consistently reduces feasibility violations in both forward- and inverse-dynamics formulations where hard-constrained methods stall (\Cref{fig:sqp_results}). While the $\ell_2$ relaxation yields only medium-accuracy feasibility, causing minor artifacts such as slight foot penetration, it nevertheless produces usable trajectories instead of solver failure. Under a $5$-hour runtime limit, \textsc{ProxSQP} variants either diverged or timed out, while iterative refinement was required to avoid immediate divergence. In contrast, \textsc{OdynSQP} converged within a few seconds.

\begin{figure}[t]
  \centering
  \hspace{5mm}
  \ifpublished
  \href{https://youtu.be/X9AgWGu5Eug}
  {
  \includegraphics[width=0.15\columnwidth]{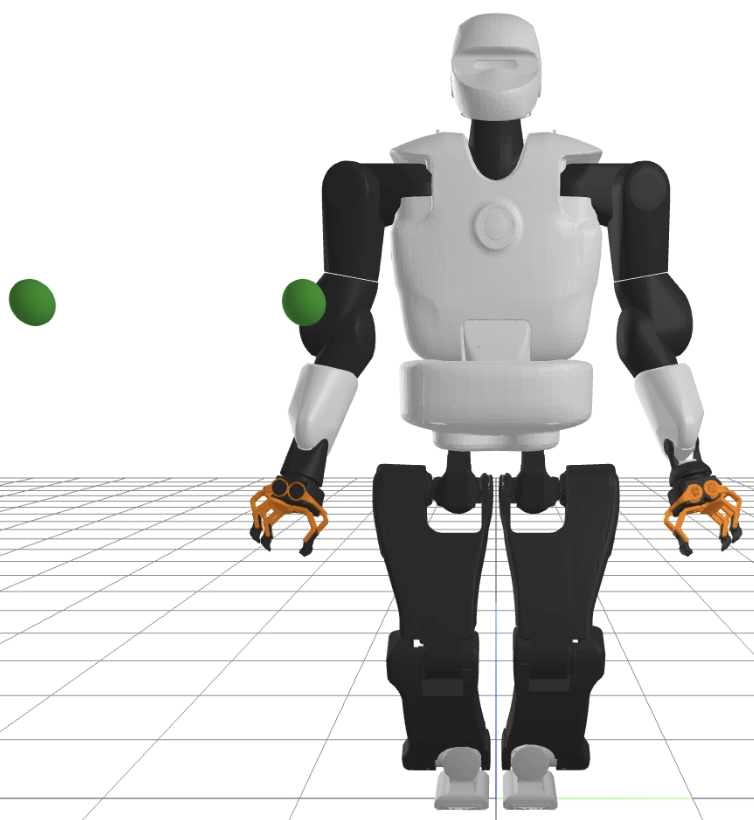}\hspace{7mm}
  \includegraphics[width=0.15\columnwidth]{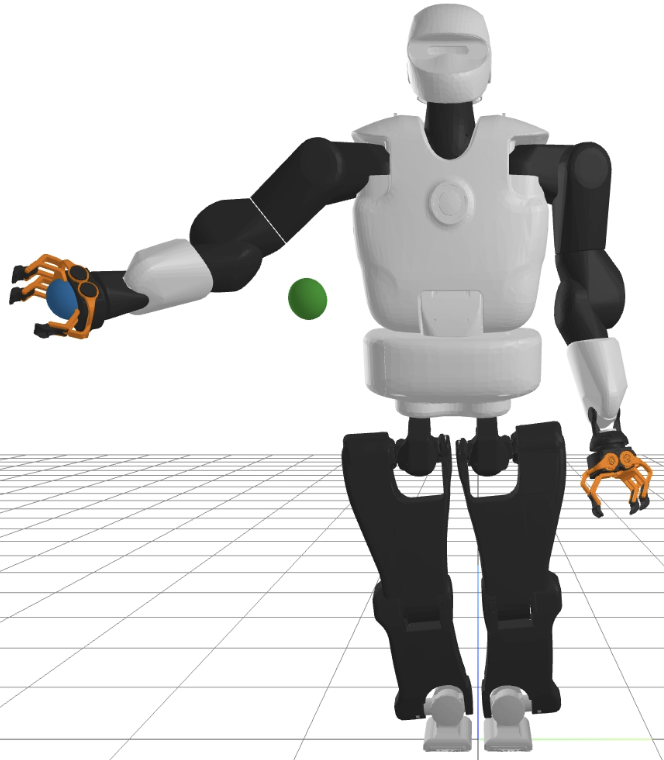}\hspace{7mm}
  \includegraphics[width=0.15\columnwidth]{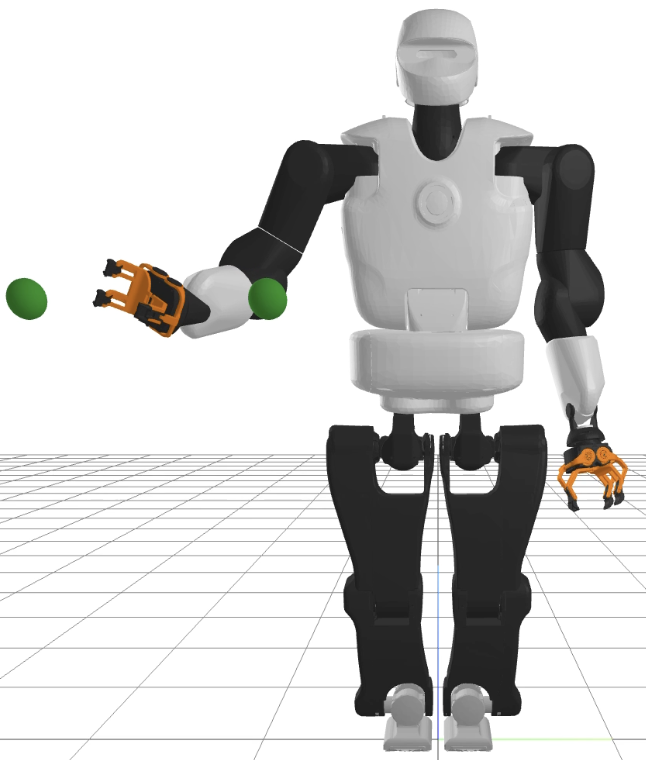}\hspace{7mm}
  \includegraphics[width=0.15\columnwidth]{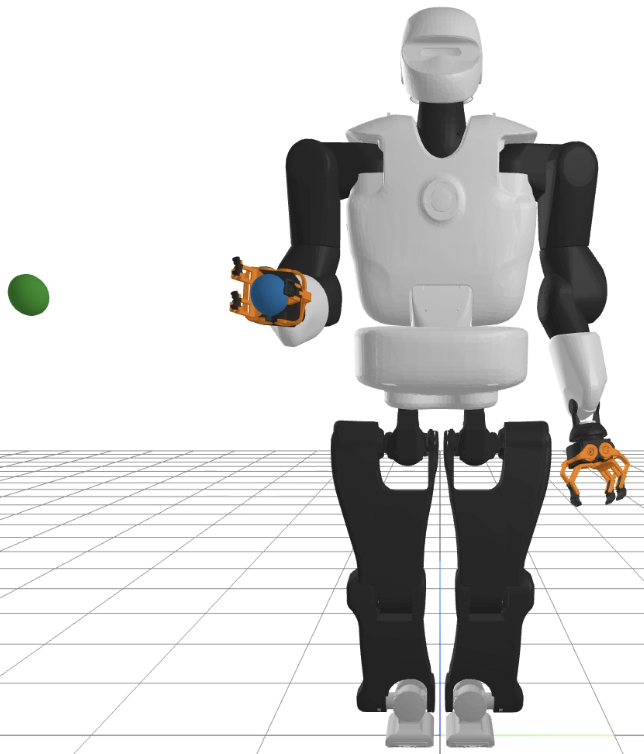}\\
  \includegraphics[width=0.38\columnwidth, trim=0 0 0 0, clip]{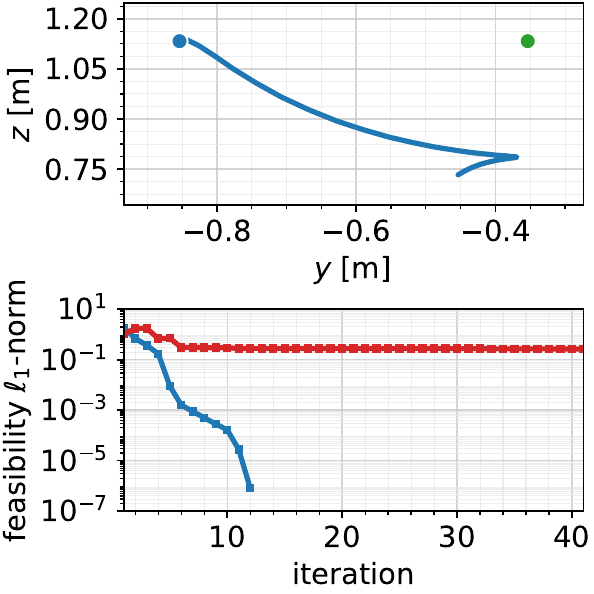}\hfill
  \includegraphics[width=0.31\columnwidth, trim=52 0 0 0, clip]{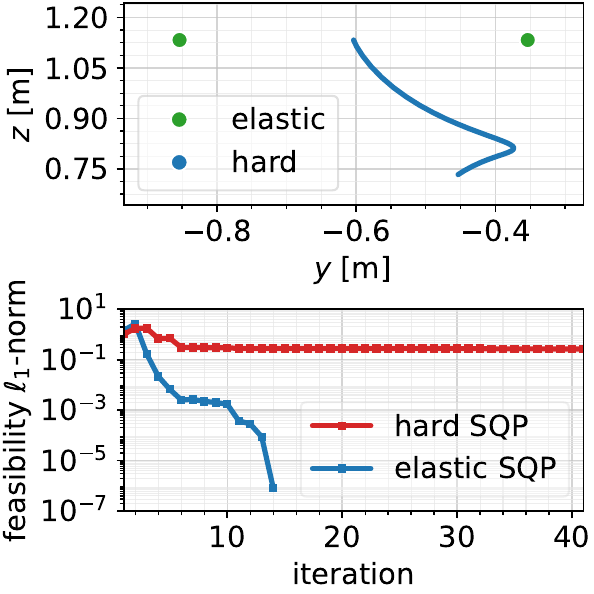}\hfill
  \includegraphics[width=0.31\columnwidth, trim=52 0 0 0, clip]{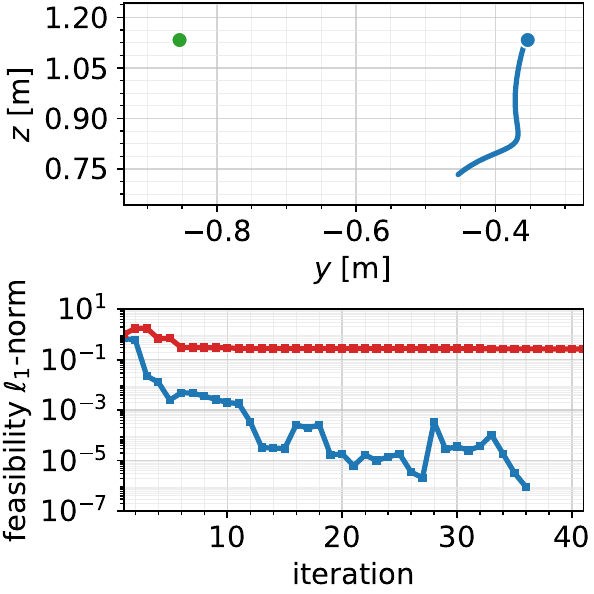}
  }
  \else
  {
  \includegraphics[width=0.16\columnwidth]{Figures/standing.png}\hspace{7mm}
  \includegraphics[width=0.16\columnwidth]{Figures/p_left.png}\hspace{7mm}
  \includegraphics[width=0.16\columnwidth]{Figures/p_mid.png}\hspace{7mm}
  \includegraphics[width=0.16\columnwidth]{Figures/p_right.png}\\
  \includegraphics[width=0.38\columnwidth, trim=0 0 0 0, clip]{Figures/humanoid_two_target_contradiction_B.pdf}\hfill
  \includegraphics[width=0.31\columnwidth, trim=52 0 0 0, clip]{Figures/humanoid_two_target_contradiction_AB.pdf}\hfill
  \includegraphics[width=0.31\columnwidth, trim=52 0 0 0, clip]{Figures/humanoid_two_target_contradiction_A.pdf}
  }
  \fi
  \vspace{-8mm}
  \caption{Infeasible trajectory optimization of Talos. Per column a constraint is
  relaxed elastically or kept hard. The \textsc{Elastic OdynSQP} converges where hard \textsc{OdynSQP} stalls, settling at the least-squares midpoint solution or the hard targets. \ifpublished Full video: \texttt{\href{https://youtu.be/X9AgWGu5Eug}{https://youtu.be/X9AgWGu5Eug}}.
  \else \fi
  }
  \label{fig:two-target}
  \vspace{-7mm}
\end{figure}

\textbf{Selective outer relaxation.} For infeasible problems with no feasible trajectory, a chosen subset of constraint categories, e.g., dynamics, path equalities, inequalities, or terminal targets, is routed to the elastic $\ell_2$ block of \textsc{Elastic Odyn}, with the rest enforced as hard constraints. This spans a continuum from full restoration, where all constraints are elastic, to minimal relaxation of only the conflicting ones, isolating infeasibility without sacrificing accuracy elsewhere. \Cref{fig:two-target} illustrates this on an infeasible Talos \reb{humanoid} task with two contradictory targets, where \textsc{Elastic OdynSQP} converges, settling at the midpoint or a hard target, while hard \textsc{OdynSQP} stalls.

\reb{\textbf{Merit and globalization.} Elastic violations enter the merit function as a squared $\ell_2$ penalty consistent with inner relaxation, while hard residuals are carried by an adaptive $\ell_1$ term,
$
\Phi_{\!E}(\mathbf{w}) = J(\mathbf{w}) + V_{\!E}(\mathbf{w}) + \upsilon_{\!H}\,c_{\!H}(\mathbf{w}),
$
with $\mathbf w := (\mathbf X, \mathbf U)$, $J=\sum_k l_k$, hard residual norm $c_{\!H}$, and elastic penalty $V_{\!E} = \tfrac{1}{2\gamma_e}\|\mathbf{h}_\mathrm{el}\|_2^2 + \tfrac{1}{2\gamma_i}\|[\mathbf{g}_\mathrm{el}]_+\|_2^2$. }

\vspace{-1mm}
\section{Conclusion}\label{sec:conclusion}
We introduced \textsc{Elastic Odyn}, a smooth $\ell_2$-based extension of a primal--dual non-interior-point QP solver that remains robust under infeasibility, degeneracy, and ill-conditioning. The method computes closest-to-feasible solutions, supports efficient warm starting, and recovers interpretable dual variables through a lightweight refinement stage. Its smooth formulation further enables stable differentiation through infeasible optimization problems, yielding a differentiable QP layer suitable for learning-based applications. Building on this foundation, \textsc{Elastic OdynSQP} extends these capabilities to nonlinear optimal control, enabling reliable convergence in the presence of inconsistent linearizations and intrinsically infeasible tasks. Together, these results establish a unified framework for optimization, simulation, control, and learning beyond the feasibility assumptions of existing methods.
\vspace*{-2mm}


\bibliographystyle{IEEEtran}
\bibliography{references}

\end{document}